\documentclass{article}
\usepackage[nonatbib,preprint]{nips_2018}

\usepackage{color}
\usepackage[utf8]{inputenc} 
\usepackage[T1]{fontenc}    
\usepackage{hyperref}       
\usepackage{url}            
\usepackage{booktabs}       
\usepackage{amsfonts}       
\usepackage{nicefrac}       
\usepackage{microtype}      
\usepackage{multirow}
\usepackage{graphicx}
\usepackage{longtable}
\graphicspath{ {images/} }

\usepackage{biblatex}
\usepackage{algorithm}
\usepackage{algorithmic}

\def\vect#1{\mbox{\boldmath $#1$}}

\usepackage{wrapfig}

\title{Study of Deep Generative Models for \\Inorganic Chemical Compositions}

\author{
  Yoshihide Sawada\\
  Panasonic Corporation\\
  \texttt{sawada.yoshihide@jp.panasonic.com} \\
  \And
  Koji Morikawa\\
  Panasonic Corporation\\
  \texttt{morikawa.koji@jp.panasonic.com} \\
  \And
  Mikiya Fujii\\
  Panasonic Corporation\\
  \texttt{fujii.mikiya001@jp.panasonic.com} \\
}

\begin{document}

\maketitle

\begin{abstract}
Generative models based on generative adversarial networks (GANs) and variational autoencoders (VAEs) have been widely studied in the fields of image generation, speech generation, and drug discovery, but, only a few studies have focused on the generation of inorganic materials. Such studies use the crystal structures of materials, but material researchers rarely store this information. Thus, we generate chemical compositions without using crystal information. We use a conditional VAE (CondVAE) and a conditional GAN (CondGAN) and show that CondGAN using the bag-of-atom representation with physical descriptors generates better compositions than other generative models. Also, we evaluate the effectiveness of the Metropolis-Hastings-based~(MH) atomic valency modification and the extrapolation performance, which is important to material discovery.
\end{abstract}

\section{Introduction}
\label{sec:intro}
In recent years, generative models based on generative adversarial networks (GANs)~\cite{goodfellow2014generative} and variational autoencoders (VAEs)~\cite{kingma2013auto} have been widely studied in the fields of image generation~\cite{brock2018large}, speech generation~\cite{oord2016wavenet}, and drug discovery~\cite{schwalbe2019generative}. Drug discovery has been rich with studies resulting from the collaboration of machine learning and materials science to generate valid organic molecules by the use of SMILES or graph representations~\cite{sanchez2018inverse,schwalbe2019generative,madhawa2019graphnvp}. However, there have been only a few studies focusing on inorganic materials~\cite{nouira2018crystalgan,hoffmann2019data,NOH2019}. 

Nouira et al.~\cite{nouira2018crystalgan} introduces geometric constraints and shows that their method can produce stable structures for particular chemicals. Also, Hoffmann et al.~\cite{hoffmann2019data} and Noh et al.~\cite{NOH2019} propose a VAE-based method to generate 3D crystal structures. Since crystal information is known to improve the performance of predictions~\cite{seko2017representation,xie2018crystal,schutt2018schnet,chen2019graph}, these methods are likely to generate valid inorganic molecules.

Physical properties are, in principle, governed by crystal structures and electron distributions, which can be explored by X-ray diffraction techniques and first-principles calculations. These methodologies, however, are not so feasible and are time-consuming, therefore only chemical compositions and observed physical properties are stored for the rapid exploration of new materials. These problems have led us to attempt the generation of chemical compositions without using crystal information.

Our goal is to achieve an innovative design, a so-called Inverse Material Design~(IMD)~\cite{jain2014perspective,sanchez2018inverse}, of inorganic materials. This design predicts promising chemical compositions when the desired properties are given, whereas the conventional development of materials is based on physical laws and physical properties predicted from chemical compositions, crystal structures, and electron distributions. To achieve this design, there are two approaches to generative models, which are reinforcement learning (RL) + unsupervised models~\cite{guimaraes2017objective,sanchez2017optimizing,de2018molgan} and conditional models~\cite{gomez2018automatic,lim2018molecular,assouel2018defactor}. RL + unsupervised models use simulators for obtaining rewards~(e.g., drug-likeness) and the unsupervised generative models generate samples to maximize the rewards. Such models have been widely used for the generation of organic molecules. However, the inorganic molecules generated by these models have to be evaluated by computationally expensive density functional theory (DFT) calculations that require crystal information. Our study focuses on conditional generative models, such as a conditional GAN~(CondGAN)~\cite{mirza2014conditional,odena2017conditional} and a conditional VAE~(CondVAE)~\cite{kingma2014semi,klys2018learning}. 

In material science, it is known that the valence of a chemical composition should become zero. However, standard CondGAN and CondVAE do not preserve this constraint. Madhawa et al.~\cite{madhawa2019graphnvp} uses a greedy beam search to explore valid graph molecules. We apply a Metropolis Hastings~(MH)-based method~\cite{bishop2006pattern} to balance the atomic valencies of generated compositions. 

To construct generative models, we use two different data representations: bag-of-atoms~\cite{bresson2019two} and bag-of-atoms with physical descriptors. Bag-of-atoms is a vector representing the number of each atom included in the composition. By using this representation, we can easily convert to the corresponding composition. The physical descriptors are characteristics computed from the compositions~\cite{xenonpy}, i.e., crystal information is not included in the descriptors.

We experiment with the Materials Project database, which has more than 60,000 inorganic compositions~\cite{jain2013commentary}, and show that CondGAN and the bag-of-atoms~\cite{bresson2019two} with physical descriptors~\cite{xenonpy} generate the best compositions. Despite the non-usage of crystal structures, CondGAN($\vect{x}^{bp}$) generates compositions around the desired property. Also, we evaluate the effectiveness of the MH-based modification and the extrapolation performance, which is an important skill for IMD~\cite{meredig2018can}.

This remainder of this paper is organized as follows. Section~\ref{sec:input} describes the bag-of-atoms and physical descriptors. Section~\ref{sec:models} describes the generative models, i.e., CondVAE and CondGAN. Section~\ref{sec:valence} describes the MH-based valency modification. Section~\ref{sec:ex_setting} and \ref{sec:ex_results} describe the setup and results, respectively, of our experiment. Section~\ref{sec:concl} discusses our conclusions and direction for future works.

\section{Representations of Chemical Compositions}
\label{sec:input}
As discussed in Sec.~\ref{sec:intro}, we use two different types of data representations: bag-of-atoms~\cite{bresson2019two} and bag-of-atoms with physical descriptors~\cite{xenonpy}. Let $\vect{x}^{b} \in \mathbb{R}^{M^{b}}$ and $\vect{x}^{p} \in \mathbb{R}^{M^{p}}$ denote vectors representing the bag-of-atoms and the physical descriptors, respectively, where $M^{b}$ and $M^{p}$ are the numbers of atoms used in the dataset and physical descriptors, respectively. 

$x^{b}_i$ represents the count of the $i$-th atom of the composition. For example, if there are 3 atoms (H, Li, O) used in the dataset and the target composition is H2O, then $\vect{x}^b = [2,0,1]^\top$. We can easily convert this representation to the composition.

Table~\ref{tbl:physical_desc} shows a list of the physical descriptors. In \cite{xenonpy}, the descriptors of 94 atoms are saved. Using these descriptors, we compute the weighted mean vector, $\vect{x}^p$~(for the example above, $\vect{x}^p = 2/3\vect{x}_H + 1/3\vect{x}_O$, where $\vect{x}_H$ and $\vect{x}_O$ are the physical descriptors' vectors of H and O, respectively). We experimented beforehand and found that this representation provided better performance than others~(e.g., weighted variance vector~\cite{xenonpy}). Note that $\vect{x}^p$ is difficult to convert to the composition, thus we use $\vect{x}^b$ and $\vect{x}^{bp} = [\vect{x}^{b\top},\vect{x}^{p\top}]^\top$ in our experiments.

\section{Generative Models}
\label{sec:models}
In this section, we describe CondVAE and CondGAN. For simplicity, we represent the input vector by $\vect{x}$.

\subsection{Conditional Variational Autoencoder (CondVAE)}
In this article, CondVAE has two networks, which are the encoder~($q_{\phi}(.)$) and decoder~($p_{\theta}(.)$)~\cite{klys2018learning}. CondVAE trains these networks by minimizing the following equation corresponding to the Evidence Lower BOund~(ELBO) of $\log p_{\theta}(\vect{x} \mid y)$:
\begin{equation}
L_{{\rm CondVAE}} = \mathbb{E}_{q_{\phi}}[\log p_{\theta} (\vect{x} \mid y, \vect{z})] - D_{KL}(q_{\phi}(\vect{z} \mid \vect{x}, y) \| p_{\theta}(\vect{z} \mid y)),
\end{equation}
where $\vect{z}$ is the latent vector, $y$ is the target property, $\mathbb{E}$ is the expectation, and $D_{KL}(.\|.)$ is the KL divergence. In this article, we set $p_{\theta}(\vect{z} \mid y) = p(\vect{z}) = \mathcal{N}(0,\vect{I})$~\cite{klys2018learning}.

\subsection{Conditional Generative Adversarial Network (CondGAN)}
\label{sub:CondGAN}
We use the auxiliary classifier GAN~(ACGAN~\cite{odena2017conditional}) as CondGAN. The main difference between the traditional CondGAN and ACGAN is the function of the discriminator, which evaluates if the input is real or fake in the former. In addition to this evaluation, the discriminator in AGAN classifies the category of the input. The loss function is as follows:
\begin{eqnarray}
L_{{\rm CondGAN}} &=& \mathbb{E}_{\tilde{{x}}} [D(\tilde{\vect{x}})] - \mathbb{E}_{{x}} [D(\vect{x})] + \lambda_1 \mathbb{E}_{\hat{{x}}} [(\|\nabla_{\hat{{x}}}D(\hat{\vect{x}}) \|_2 - 1)^2] \nonumber\\
&& + \lambda_2 \mathbb{E}_{\tilde{{x}}} [(y-P(\tilde{\vect{x}}))^2] + \lambda_3 \mathbb{E}_{{x}} [(y-P(\vect{x}))^2],
\end{eqnarray}
where $D(.)$ is the output of the real/fake evaluation, $P(.)$ is the predicted output, and $\tilde{\vect{x}}$ is the output of the generator~($\tilde{\vect{x}} = G(\vect{z})$), $\hat{\vect{x}} = \epsilon \vect{x} + (1-\epsilon)\tilde{\vect{x}}$ ($\epsilon \sim U[-1,1]$, where $U[-1,1]$ is the uniform distribution ranging from $-1$ to $1$). $\lambda_1, \lambda_2$, and $\lambda_3$ are hyperparameters. From the first to the third terms are the Wasserstein losses~\cite{gulrajani2017improved}, which are known to achieve stable training, whereas the other terms are the auxiliary losses~\cite{odena2017conditional}. It should be noted that we use the least square error instead of the cross-entropy loss because the property is not categorical.

\section{Valency-based Vector Modification}
\label{sec:valence}
In material science, it is known that the valence of a composition has to become zero. However, the generative models described in the previous section may violate this condition because there is no constraint. We solve this problem by applying an Metropolis-Hastings~(MH) method, which is one of the Markov Chain Monte Carlo~\cite{bishop2006pattern}. MH-based method modifies $\vect{x}^b$ according to the following acceptance probability $\alpha$:
\begin{eqnarray}
\alpha(\vect{x}^b_{{\rm new}} \mid \vect{x}^b) = \min \left( 1, \frac{q(\vect{x}^b \mid \vect{x}^b_{{\rm new}}) \pi(\vect{x}^b_{{\rm new}})}{q(\vect{x}^b_{{\rm new}} \mid \vect{x}^b) \pi(\vect{x}^b) }\right).
\end{eqnarray}
We use the proposal distribution $q(. \mid .)$ as a Gaussian and $\pi(.)$ as follows:
\begin{eqnarray}
\pi(\vect{x}^b) = \max_{v} \left( \exp(-(\vect{v}^\top\vect{x}^b)^\top (\vect{v}^\top\vect{x}^b)) \right), 
\end{eqnarray}
where $\vect{v}$ is the valence vector that $v_i$ corresponds to the valence of the $i$-th atom. Since there are atoms with multiple valences, we use the maximum value as $\pi(.)$. 

It is known that the MH method is inefficient when $\vect{x}^b$ is a high dimension~\cite{bishop2006pattern}. Thus, we compress $\vect{x}^b$ so that it satisfied only the variable $x^b_i > TH$. Also, to accelerate the MH search, we sample $N$ vectors simultaneously and use $\max(\alpha_j), (j=1,2,\cdots,N)$ as the acceptance probability. If $\pi(.) = 1$, we stop and convert $\vect{x}^b_{\rm new}$ to the composition.

\section{Experimental Setup}
\label{sec:ex_setting}
Before presenting the experimental results, we explain the dataset, network architecture, and hyperparameters. For this implementation, we use Python libraries such as Pymatgen~\cite{ong2013python}, Xenonpy~\cite{xenonpy}, and TensorFlow~\cite{tensorflow}~\footnote{Sample source code for training CondGAN($\vect{x}^{bp}$) is here: https://github.com/yoshihidesawada/CompGAN}.

\subsection{Dataset}
For the experiment, we use 69,640 compositions with their formation energies~[eV/atom] in the Materials Project~\cite{jain2013commentary}. Figure~\ref{fig:formation_hist} shows a histogram of the formation energies, which range from $-4.5$~[eV/atom] to $4.4$~[eV/atom] (almost $< 0$ [eV/atom]) and have a bimodal distribution. After removing duplicate compositions, we randomly select 44,040 training data and 5,506 test data. In the evaluation, we generate 256 compositions for each test property. 

\subsection{Network Architecture and Hyperparameters}
The number of hidden layers in the generator and the decoder is two while the dimensions of the first and second layers are 60 and 30, respectively. The discriminator and the encoder have the same number of layers and their dimensions are the inverse of the generator/decoder. All models use the fully connected ReLU layer and the optimization is Adam. $\vect{x}^b$ and $\vect{x}^{bp}$ are normalized by $x_i = (x_i-{\sf min}(x_i))/({\sf max}(x_i)-{\sf min}(x_i))$, where ${\sf max}(x_i)$ and ${\sf min}(x_i)$ are the maximum and minimum, respectively, of the $i$-th variable. Also, $M^b=89$, $M^p=58$, $\vect{z} \in \mathbb{R}^{10}$, $\lambda_1=10$, and $\lambda_2=\lambda_3=1$. The batch and epoch sizes are 256 and 50,000, respectively. To convert the compositions, we use only atoms satisfying $\vect{x}^b > TH~(=0.03)$, $N=100$, and normalize their variables to 1. Note that these parameters are empirically determined.

\section{Experimental Results}
\label{sec:ex_results}
\subsection{Generated Compositions}
Table~\ref{tbl:generated_comps} shows the generated compositions of each model. Note that the compositions shown in this section are converted to $\vect{x}^b$ without the use of the MH-based method. Thus, the generated compositions include non-valid molecules~(e.g., Li0.27O0.72Cr0.01 in Table~\ref{tbl:generated_comps}). 

As shown in table~\ref{tbl:generated_comps}, CondGAN($\vect{x}^{bp}$) generates certain compositions but not others. These generations may have been due to posterior collapse~\cite{razavi2019preventing} and mode collapse~\cite{goodfellow2016nips}. These collapses are phenomena such that $D_{KL}(q_{\phi}(\vect{z} \mid \vect{x}) \|  p(\vect{z})) = 0$ and the outputs of GANs degenerates to the mode. However, it is not yet clear why concatenating $\vect{x}^p$ mitigates only CondGAN. A deeper analysis will constitute a future study.

Table~\ref{tbl:rmse} shows the mean absolute error~(MAE), root mean squared error~(RMSE), and distance $d(\tilde{\vect{x}}, {\sf near}(\tilde{\vect{x}}))/{\sf dim}(\tilde{\vect{x}})$, where $\tilde{\vect{x}}$ is the generated vector, ${\sf near}(\tilde{\vect{x}})$ is the nearest vector, $d(.,.)$ is the Euclidean distance, and ${\sf dim}(\tilde{\vect{x}})$ is the dimension of $\tilde{\vect{x}}$. MAE and RMSE are computed between the input property $y$ and the nearest property ${\sf near}(y)$ corresponding to the property of ${\sf near}(\tilde{\vect{x}})$. For references, we also evaluate the performances of CondVAE($\vect{x}^{bp}$) and CondGAN($\vect{x}^{bp}$) using only $\vect{x}^b$. As shown in table~\ref{tbl:rmse}, $\tilde{\vect{x}}$ that is generated by CondGAN($\vect{x}^{bp}$) is the closest to the real vector ${\sf near}(\tilde{\vect{x}})$ with the desired property. 

\begin{figure}[t]
\centering
\includegraphics[width=0.465\linewidth]{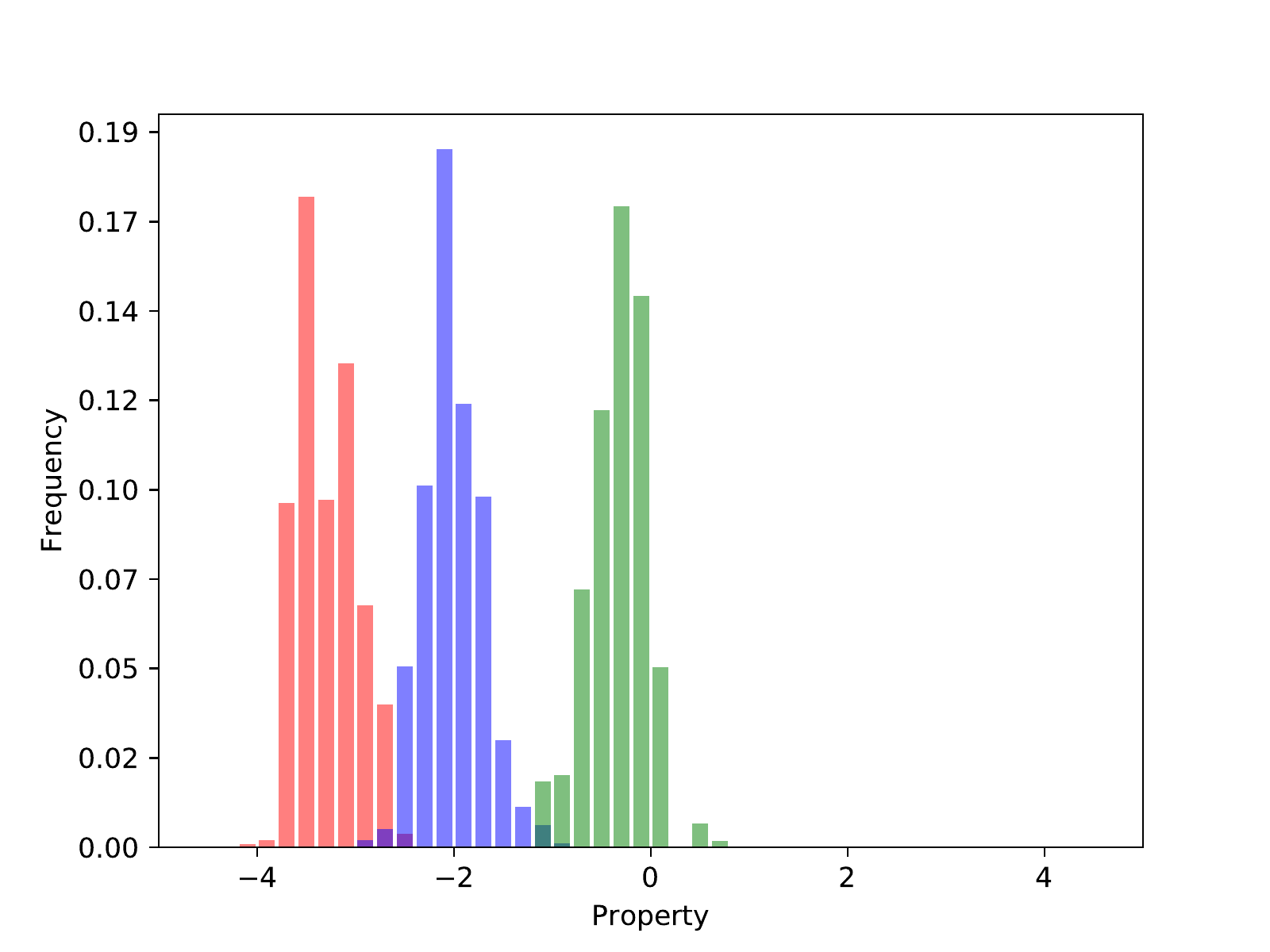}
\includegraphics[width=0.425\linewidth]{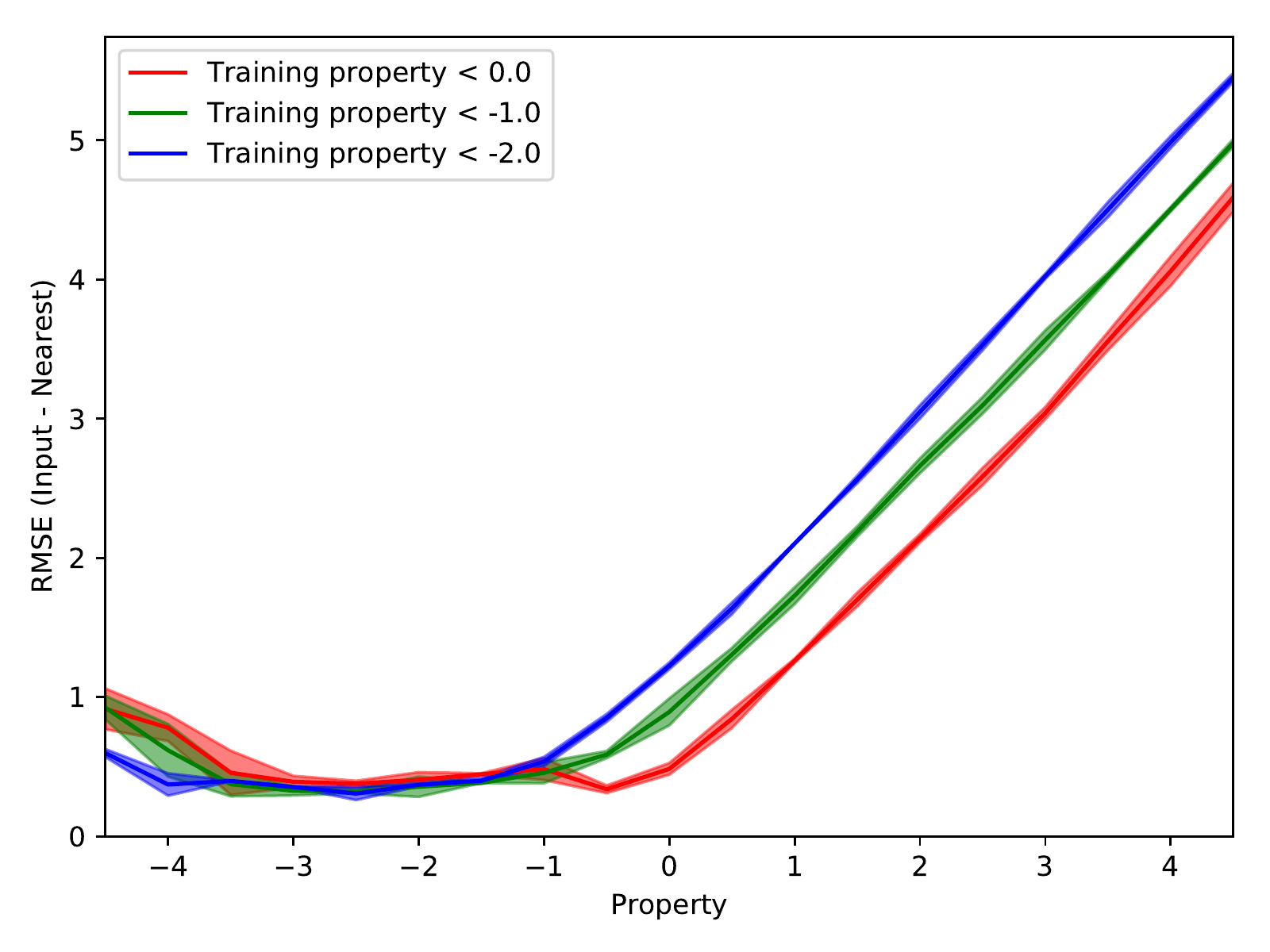}\\
~~~~~~~(A)~~~~~~~~~~~~~~~~~~~~~~~~~~~~~~~~~~~~~~~~~~~~~~~~~~~~~~~~~~~~~~~~~~(B)
\caption{(A) Histogram of the nearest property of the generated compositions. From left to right, the desired properties~(input) are $-3.24$~(red), $-2.03$~(blue), and $-0.10$~(green). These property values are not included in the training dataset. (B) Evaluation of the extrapolation. To evaluate the relationship between the extrapolation and the range of training data, we use three types of training data, $-4.5 \leq y \leq 0.0$~(red), $-4.5 \leq y \leq -1.0$~(green), and $-4.5 \leq y \leq -2.0$~(blue).}
\label{fig:results}
\end{figure}

Figure~\ref{fig:results}~(A) and table~\ref{tbl:near_comp} show the histogram of ${\sf near}(y)$ by CondGAN($\vect{x}^{bp}$), as well as examples of the generated and nearest compositions with input, predicted, and nearest properties. These results indicate that CondGAN($\vect{x}^{bp}$) generates compositions around the desired property despite the non-usage of crystal structures.

\begin{table}[t]
\caption{Examples of generated inorganic compositions.}
\label{tbl:generated_comps}
\centering
\begin{tabular}{|c|c|}
\hline
CondVAE($\vect{x}^b$) & CondVAE($\vect{x}^{bp}$)\\
\hline \hline
Li0.09O0.65F0.08S0.09Se0.09 & H0.09Li0.06O0.68F0.09S0.08\\
Li0.09O0.65F0.08S0.09Se0.09	& H0.09Li0.06O0.68F0.09S0.08\\
Li0.09O0.65F0.08S0.09Se0.09 & H0.09Li0.06O0.68F0.09S0.08\\
Li0.09O0.65F0.08S0.09Se0.09 & H0.09Li0.06O0.68F0.09S0.08\\
Li0.09O0.65F0.08S0.09Se0.09 & H0.09Li0.06O0.68F0.09S0.08\\
Li0.09O0.65F0.08S0.09Se0.09 & H0.09Li0.06O0.68F0.09S0.08\\
Li0.09O0.65F0.08S0.09Se0.09 & H0.09Li0.06O0.68F0.09S0.08\\
Li0.09O0.66F0.08S0.09Se0.08 & H0.09Li0.06O0.68F0.09S0.08\\
Li0.09O0.66F0.08S0.08Se0.09 & H0.09Li0.06O0.68F0.09S0.08\\
\hline
\hline
CondGAN($\vect{x}^{b}$) & CondGAN($\vect{x}^{bp}$)\\
\hline \hline
F0.60As0.40 & Li0.29O0.58Mn0.13\\
F0.92K0.04As0.04 & Li0.27O0.72Cr0.01\\
O0.04F0.88K0.04Xe0.04 & S0.01Pr0.59Sm0.40\\
O0.04F0.84K0.04As0.04Xe0.04 & Li0.22O0.59Cr0.11Mn0.08\\
O0.05F0.83K0.04As0.04Xe0.04 & S0.03Pr0.72Sm0.19U0.06\\
O0.06F0.81K0.05As0.04Xe0.04 & O0.62Na0.09Ta0.01Au0.28\\
O0.06F0.81K0.05As0.04Xe0.04 & S0.01Pr0.60Sm0.01Dy0.20U0.18\\
O0.06F0.81K0.05As0.04Xe0.04 & H0.04C0.01O0.71Na0.24\\
O0.06F0.81K0.05As0.04Xe0.04 & O0.85Ca0.11Ta0.01W0.02Au0.01\\
\hline
\end{tabular}
\end{table}

\begin{table}[t]
\caption{Comparison of models. We compute MSE/RMSE between ${\sf near}(y)$ and $y$, as well as the distance $d(\tilde{\vect{x}}, {\sf near}(\tilde{\vect{x}}))/{\sf dim}(\tilde{\vect{x}})$. We execute the random training/test data selection three times to compute~(mean$\pm$std).}
\centering
\begin{tabular}{|c|c|c|c|c|}
\hline
\multicolumn{2}{|c|}{} & MAE & RMSE & Distance$\times10^3$\\
\hline \hline
\multicolumn{2}{|c|}{CondVAE($\vect{x}^b$)} & $0.53 \pm 0.11$ & $0.64\pm0.11$ & $4.45\pm0.07$\\
\hline
\multicolumn{2}{|c|}{CondGAN($\vect{x}^{b}$)} & $0.53 \pm 0.05$ & $0.68\pm0.08$ & $4.97\pm0.43$\\
\hline
\multirow{2}{*}{CondVAE($\vect{x}^{bp}$)} & $\vect{x}^b$ & $0.46 \pm 0.13$ & $0.60\pm0.15$ & $4.67\pm0.36$\\
 & $\vect{x}^{bp}$ & $0.57 \pm0.11$ & $0.74\pm0.12$ & $3.13\pm0.06$\\
\hline
\multirow{2}{*}{CondGAN($\vect{x}^{bp}$)} & $\vect{x}^b$ & $0.39\pm0.01$ & $0.52\pm0.01$ & $3.81\pm0.11$\\
 & $\vect{x}^{bp}$ & $0.37\pm0.03$ & $0.46\pm0.01$ & $2.78\pm0.07$\\
\hline
\end{tabular}
\label{tbl:rmse}
\end{table}

\begin{table}[t]
\caption{Examples of generated compositions, predicted property $P(\tilde{\vect{x}})$, nearest composition, and property of nearest composition ${\sf near}(y)$. Note that the composition in parentheses of the nearest composition's line is the normalized composition.}
\centering
\small
\begin{tabular}{|c|c|c|c|c|}
\hline
$y$ & Generated composition & $P(\tilde{\vect{x}})$ & Nearest composition & ${\sf near}(y)$\\
\hline \hline
$-3.25$ & O0.61Si0.01Ba0.38 & $-3.25$ & Ba9Sc2(SiO4)6 (O0.58Si0.15Sc0.05Ba0.22) & $-3.46$\\
$-3.25$ & B0.04O0.65Ca0.30Sm0.01 & $-3.16$ & Ca5B3O9F (B0.17O0.50F0.06Ca0.28) & $-3.40$\\
$-2.03$ & Li0.29O0.58Mn0.13 & $-2.02$ & Li4Mn5FeO12 (Li0.18O0.55Mn0.23Fe0.05) & $-2.00$\\
$-2.03$ & Li0.27O0.72Cr0.01 & $-2.00$ & LiCr10O15 (Li0.04O0.58Cr0.38) & $-2.23$\\
$-1.27$ & C0.02O0.98 & $-1.24$ & CrC5SO7 (C0.36O0.50S0.07Cr0.07) & $-1.03$\\
$-1.27$ & Na0.80S0.04Cu0.16 & $-1.27$ & Na2S (Na0.67S0.33) & $-1.26$\\
$-1.05$ & O0.34S0.05Zn0.61 & $-1.02$ & Zn3CdS4 (S0.50Zn0.38Cd0.12) & $-1.08$\\
$-1.05$ & Ge0.02Cd0.10Eu0.02Er0.86 & $-1.13$ & ErGe (Ge0.50Er0.50) & $-0.85$\\
$-0.10$ & Mg0.63Co0.01Sr0.10Dy0.26 & $-0.07$ & Dy5Mg24 (Mg0.83Dy0.17) & $-0.04$\\
$-0.10$ & Zn0.10Cd0.89Hg0.01 & $-0.08$ & EuCd11 (Cd0.92Eu0.08) & $-0.13$\\
\hline
\end{tabular}
\label{tbl:near_comp}
\end{table}

\subsection{Predicted Performance of Discriminator}
As described in Sec.\ref{sub:CondGAN}, the discriminator of our CondGAN is able to predict not only real/fake but also property. Thus, we evaluate the performance of $P(.)$.

Table~\ref{tlb:discriminator} shows the MAE/RMSE of each CondGAN model. For comparison, we also show the performance of DNN, which has the same hyperparameters as the discriminator. CondGAN($\vect{x}^{bp}$) provides better predictions than CondGAN($\vect{x}^{b}$). Also, CondGAN($\vect{x}^{bp}$) has a competitive performance to the standard DNN.

We speculate that $P(.)$ relates to the differences in the performances of the generative models. Thus, checking $P(.)$ may address the difficulty of checking the generative performances.

\begin{table}[t]
\caption{Comparison of the performances of $P(.)$. (mean$\pm$std) is computed by the same random selection shown in table~\ref{tbl:rmse}.}
\centering
\begin{tabular}{|c|c|c|c|}
\hline
 & CondGAN($\vect{x}^b$) & CondGAN($\vect{x}^{bp}$) & DNN\\
\hline
\hline
 MAE & $0.20\pm0.00$ & $0.12\pm0.00$ & $0.10\pm0.01$\\
 RMSE & $0.30\pm0.01$ & $0.20\pm0.01$ & $0.22\pm0.03$\\
\hline
\end{tabular}
\label{tlb:discriminator}
\end{table}

\subsection{Valence Check and Modifications of Compositions}
We evaluate the MH-based valency modification. Tables~\ref{tbl:rmse_mh} and \ref{tbl:valence} show the evaluation performances of the MH-based modification and examples of the modified compositions. To compute these values, 10 properties ($-4.49, -3.76, -3.45, -3.01, -2.50, -2.03. -1.52, -1.07, -0.48$, and $-0.01$) are fed into CondGAN($\vect{x}^{bp}$), which generates 256 samples per property. 

As shown in Table~\ref{tbl:rmse_mh}, the MH-based modification reduces the violation of valency and the distance to the nearest composition while maintaining the performances of MAE/RMSE. However, we find compositions that cannot balance the atomic valency~(e.g., Gd0.06Tb0.74Ac0.19 and Al0.10Ho0.90 in table~\ref{tbl:valence}) because the MH-based method is handled with only $\vect{x}^{b}>TH$. Intuitively speaking, applying the MH-based method to all dimensions can solve this problem, however such a solution is time-consuming. Thus, we will plan to constrain the atomic valency during the training phase of the generative model.

\begin{table}[t]
\caption{Evaluation of the MH-based valency modification. Raw represents the generated outputs of CondGAN($\vect{x}^{bp}$) and Modified represents the results of the MH-based modification.}
\centering
\begin{tabular}{|c|c|c|c|c|}
\hline
& $\mid {\rm Valence} \mid$ & MAE & RMSE & Distance$\times10^3$\\
\hline \hline
Raw & $1.02 \pm 0.52$ & $0.50 \pm 0.19$ & $0.70\pm0.26$ & $2.49\pm0.78$\\
\hline
Modified & $0.19 \pm 0.24$ & $0.53 \pm 0.16$ & $0.70\pm0.18$ & $1.63\pm0.33$\\
\hline
\end{tabular}
\label{tbl:rmse_mh}
\end{table}

\begin{table}[t]
\caption{Examples of compositions modified by the MH-based algorithm.}
\label{tbl:valence}
\centering
\begin{tabular}{|c|c|}
\hline
Generated composition & Modified generated composition\\
\hline \hline
O0.65Y0.11Ba0.04La0.20 & O0.58Y0.12Ba0.12La0.18\\
O0.74La0.26 & O0.60La0.40\\
O0.57La0.43 & O0.60La0.40\\
O0.10F0.58Mg0.12Ba0.06Pr0.14 & O0.10F0.58Mg0.12Ba0.06Pr0.14\\
Li0.07O0.53F0.09Co0.31 & O0.50Co0.50\\
S0.84Ba0.16 & S0.50Ba0.50\\
Li0.84O0.06Ni0.10 & Li0.56O0.36Ni0.08\\
K0.06Se0.65In0.11Cs0.18 & Se0.38In0.07Cs0.55\\
Ce0.25Gd0.15Tb0.11Ir0.16Ac0.27Th0.06 & Gd0.06Tb0.74Ac0.19\\
Al0.85Ho0.15 & Al0.10Ho0.90\\
\hline
\end{tabular}
\end{table}

\subsection{Extrapolation}
Extrapolation is an important skill for finding compositions with desirable properties~\cite{meredig2018can}. However, previous results cannot show the extrapolation performance because these evaluate only interpolated results. hence, we evaluate the extrapolation performance of CondGAN($\vect{x}^{bp}$) by changing the selection of training/test data.

Fig.~\ref{fig:results}~(B) shows the RMSE between $y$ and ${\sf near}(y)$. The generative performance becomes worse as the distance from the range of $y$ increases. Also, the extrapolation performance becomes better when the range of $y$ increases. This result implies that the functions of collecting and evaluating the outside data are indispensable. In the field of organic molecules, this function is implemented by RL techniques~\cite{sanchez2017optimizing,de2018molgan}. However, as described in Sec.~\ref{sec:intro}, achieving the reward efficiently is difficult for the generation of inorganic molecules. We intend to address this problem in a future study.

\section{Conclusion and Future Works}
\label{sec:concl}
In this study, we attempt to generate inorganic compositions without using crystal information. We construct four types of generative models and find that CondGAN($\vect{x}^{bp}$) perform the best. Also, we apply the MH-based modifications and show that this method is effective for balancing the atomic valency. Furthermore, we evaluate the extrapolation performance of CondGAN($\vect{x}^{bp}$) and confirm that the present generator can not generate chemical compositions while holding physical properties that lay outside the range of the properties included in the training data. The simplest way to overcome the difficulty is by collecting new data, including desired physical properties, in a broader range. 

In the case of organic molecules, Al\'an Aspuru-Guzik et al.~\cite{guimaraes2017objective,sanchez2017optimizing} proposes an RL-based method to generate organic molecules by holding an optimized physical property even if that property lies outside of the initial training data. Such techniques for generators to optimize the physical properties of inorganic materials have not been achieved as far as we know and should be addressed in the future. Also, there should be improvements to the generative models and an analysis of why CondGAN($\vect{x}^{bp}$) works better.

\section*{Acknowledgments}
We would like to thank Mila members for their useful input and discussion.

\printbibliography

\clearpage
\section*{Appendix}

\setcounter{section}{1}
\setcounter{table}{0}
\setcounter{figure}{0}
\def\thesection{\Alph{section}}
\def\thetable{\Alph{table}}
\def\thefigure{\Alph{figure}}

\begin{center}
\begin{longtable}[here]{|c|}
\caption{List of physical descriptors. These 58 descriptors of 94 atoms~(from H to Pu) have been saved in the Xenonpy library~\cite{xenonpy}.}
\label{tbl:physical_desc}
\\
\hline
Descriptors\\
\hline \hline
\endfirsthead
Period in the periodic table \\ \hline Number of protons found in the nucleus of an atom\\
\hline 
Atom number in mendeleev's periodic table \\ \hline Atomic radius\\
\hline
Atomic radius by Rahm et al \\ \hline Atomic volume\\
\hline
The mass of an atom \\ \hline Atom volume in ICSD database\\
\hline
physicalal dimension of unit cells in a crystal lattice \\ \hline Van der Waals radius\\
\hline
Van der Waals radius according to Alvarez \\ \hline Van der Waals radius according to Batsanov\\
\hline
Van der Waals radius according to Bondi \\ \hline Van der Waals radius from the DREIDING FF\\
\hline
Van der Waals radius from the MM3 FF \\ \hline Van der Waals radius according to Rowland and Taylor\\
\hline
Van der Waals radius according to Truhlar \\ \hline Van der Waals radius from the UFF\\
\hline
Covalent radius by Bragg \\ \hline Covalent radius by Cerdero et al\\
\hline
Single bond covalent radius by Pyykko et al \\ \hline Double bond covalent radius by Pyykko et al\\
\hline
Triple bond covalent radius by Pyykko et al \\ \hline Covalent radius by Slater\\
\hline
$C_6$ dispersion coefficient in a.u \\ \hline $C_6$ dispersion coefficient in a.u\\
\hline
Density at 295K \\ \hline Proton affinity\\
\hline
Dipole polarizability \\ \hline Electron affinity\\
\hline
Tendency of an atom to attract a shared pair of electrons \\ \hline Allen's scale of electronegativity\\
\hline
Ghosh's scale of electronegativity \\ \hline Mulliken's scale of electronegativity\\
\hline
DFT bandgap energy of T=0K ground state \\ \hline Estimated FCC lattice parameter based on the DFT volume\\
\hline
Estimated BCC lattice parameter based on the DFT volume \\ \hline Estimated FCC lattice parameter based on the DFT volume\\
\hline
DFT magnetic momenet of T=0K ground state \\ \hline DFT volume per atom of T=0K ground state\\
\hline
Herfindahl-Hirschman Index (HHI) production values \\ \hline Herfindahl-Hirschman Index (HHI) reserves values\\
\hline
Specific heat at 20oC \\ \hline Gas basicity\\
\hline
First ionisation energy \\ \hline Fusion heat\\
\hline
Heat of formation \\ \hline Mass specific heat capacity\\
\hline
Molar specific heat capacity \\ \hline Evaporation heat\\
\hline
Coefficient of linear expansion \\ \hline Boiling temperature\\
\hline
Brinell Hardness Number\\ \hline Bulk modulus\\
\hline
Melting point \\ \hline Single-bond metallic radius\\
\hline
Metallic radius with 12 nearest neighbors \\ \hline Thermal conductivity at 25 C\\
\hline
Speed of sound \\ \hline Value of Vickers hardness test\\
\hline
Ability to form instantaneous dipoles \\ \hline Young's modulus\\
\hline
Poisson's ratio \\ \hline Molar volume\\
\hline
Total unfilled electron \\ \hline Total valance electron\\
\hline
Unfilled electron in d shell \\ \hline Valance electron in d shell\\
\hline
Unfilled electron in f shell \\ \hline Valance electron in f shell\\
\hline
Unfilled electron in p shell \\ \hline Valance electron in p shell\\
\hline
Unfilled electron in s shell \\ \hline Valance electron in s shell\\
\hline
\end{longtable}
\end{center}

\begin{figure}[h]
\centering
\includegraphics[width=0.5\linewidth]{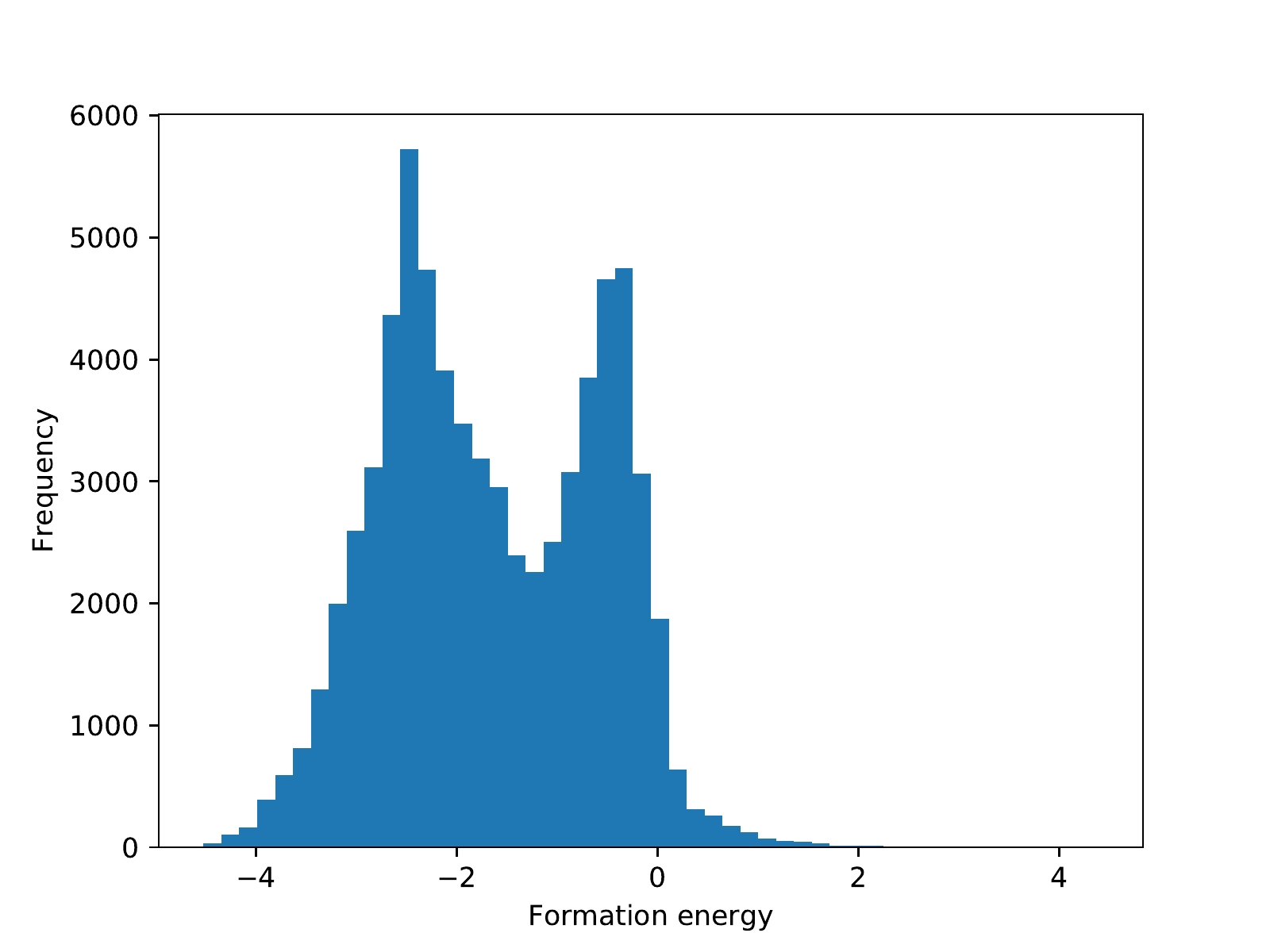}
\caption{Histogram of the formation energies~[eV/atom] in the Materials Project.}
\label{fig:formation_hist}
\end{figure}

\end{document}